\documentclass[preprint,12pt,authoryear]{elsarticle}

\usepackage{amsmath,amsfonts,amssymb,amsthm}
\usepackage{booktabs}
\usepackage{graphicx}
\usepackage{hyperref}
\usepackage{url}

\usepackage{amsmath,amsfonts,bm}

\def\eqref#1{equation~\ref{#1}}

\def\1{\bm{1}}

\DeclareMathAlphabet{\mathsfit}{\encodingdefault}{\sfdefault}{m}{sl}
\SetMathAlphabet{\mathsfit}{bold}{\encodingdefault}{\sfdefault}{bx}{n}

\newtheorem{definition}{Definition}

\graphicspath{{figures/}}

\newcommand{\Nmax}{N_{\max}}
\newcommand{\ppl}{\mathrm{ppl}}

\begin{document}

\begin{frontmatter}

\title{Anti-Periodic Positional Encoding: M\"obius Boundary Conditions Make In-Context Retrieval Reliable}

\author[aff1]{Ji Ho Bae\corref{cor1}}
\ead{jihobae@snu.ac.kr}
\cortext[cor1]{Corresponding author.}
\affiliation[aff1]{organization={JRTI},
                   city={Seoul},
                   postcode={05855},
                   country={Republic of Korea}}

\begin{abstract}
M\"obius RoPE is a rotary positional encoding built on the anti-periodic frequency ladder $\theta_i=\pi(2i+1)/N$: every rotation plane advances by an odd multiple of $\pi$ across the training context, so the positional holonomy is $-1$ and the two ends of the sequence are deterministically coupled through a closed-form Dirichlet ``dipole''; to our knowledge this is the first anti-periodic boundary condition in positional encoding. We verify the theory numerically to ${\sim}10^{-6}$ and pretrain 48 models spanning six 160M-class and three 410M-class arms (2B FineWeb-Edu tokens each; the hybrid arm puts M\"obius frequencies on 25\% of heads). Hybrid perplexity is unchanged (29.66 vs.\ 29.72), but needle-in-a-haystack retrieval becomes reliable: $90.3\pm5.7\%$ versus $63.3\pm31.4\%$ at context 512 ($n{=}6$ seeds), observed worst seed 86\% versus 14\%, robust variance tests $p=0.013$--$0.029$ (unadjusted), recurring at 410M (Levene $p=0.040$). Matched controls isolate the mechanism: an aperiodic ladder in the same frequency band reproduces none of the effect, and a periodic (holonomy $+1$) ladder only a fraction. Swapping trained models' frequency table back to standard RoPE (weights frozen) collapses retrieval, with damage concentrated on far needles: trained models depend on this long-range geometry. A NoPE arm is even more reliable at short context but pays a 13\% perplexity tax and extrapolates worst; only the anti-periodic hybrid pairs baseline perplexity with a high reliability floor. The effect is scoped to single-needle retrieval within the training window; a one-line frequency swap thus provides zero-cost insurance against the retrieval seed lottery.
\end{abstract}

\begin{keyword}
positional encoding \sep rotary position embedding \sep in-context retrieval \sep training stability \sep anti-periodic boundary condition \sep language model pretraining
\end{keyword}

\end{frontmatter}

\section{Introduction}

Whether a pretrained language model can retrieve a fact planted earlier in its context is one of the most consequential capabilities that emerges (or fails to emerge) during pretraining. We find experimentally that at small scale this capability is subject to a severe \textbf{seed lottery}: six 160M-class (162M-parameter) models pretrained identically with standard RoPE, differing only in random seed, span needle-in-a-haystack (NIAH) accuracies from 14\% to 98\% at a context of 512 tokens (\S\ref{sec:main160}). Perplexity gives no warning: the retrieval-failing seeds are not worse language models (correlation between perplexity and retrieval across the 12 RoPE-vs-hybrid models is only $+0.206$; across the study's 30-model five-arm cohort the association even runs against quality, \S\ref{sec:controls}). A practitioner who trains one model and sees healthy loss curves has no way to know whether the resulting model retrieves reliably or not.

This paper proposes a deterministic geometric intervention that largely eliminates this lottery. Standard RoPE \citep{su2024roformer} rotates each two-dimensional query/key plane of a $d_h$-dimensional attention head at frequency $\theta_i = 10000^{-2i/d_h}$; these frequencies share no short common period, so at large relative distance the per-plane phases scramble and the positional inner product wanders erratically. We ask: what if positions lived on a circle with an \textbf{anti-periodic} boundary condition (the condition obeyed by fermionic fields, whose allowed modes are the half-integer harmonics)? Setting $\theta_i = \pi(2i+1)/N$, every rotation plane advances by an odd multiple of $\pi$ after $N$ positions, so the composite rotation after one full traversal is exactly $-I$: the positional holonomy is $-1$. This has three closed-form consequences (\S\ref{sec:theory}): (i) the position-only attention profile is a Dirichlet kernel that is essentially silent over the interior of the sequence (largest interior response ${\le}2.2\%$ of the endpoint response at $d_h{=}64$); (ii) the two ends of the context are deterministically coupled with a sign flip, an end-to-end dipole (informally, a ``wormhole'' between the sequence boundaries); and (iii) under length extrapolation the positional coefficient remains bounded and exactly $2N$-periodic, in contrast to standard RoPE's aperiodic phase drift. We verify each prediction numerically to ${\sim}10^{-6}$.

Anti-periodicity is a strong prior, so we do not impose it everywhere. Our \textbf{hybrid} scheme gives M\"obius frequencies (with $N$ fixed to the training context, 2048) to 25\% of attention heads and standard RoPE to the rest, adding zero parameters and zero FLOPs: it is literally a different constant table of sines and cosines for a quarter of the heads.

\paragraph{Contributions.}
\begin{enumerate}
\item \textbf{A new class of positional encoding.} We introduce anti-periodic boundary conditions into positional encoding, with closed-form theory (Dirichlet-kernel attention signature, $-1$ holonomy, dipole structure, bounded $2N$-periodic extrapolation), verified numerically to ${\sim}10^{-6}$ (\S\ref{sec:theory}, \S\ref{sec:numerics}).
\item \textbf{Main empirical result: retrieval reliability.} In $n{=}6$-seed pretraining runs at 160M/2B tokens, hybrid M\"obius RoPE matches standard RoPE's perplexity (29.66 vs.\ 29.72) while raising mean NIAH accuracy at every in-training length (suggestive, not individually significant; min $p=0.09$) and, more importantly, collapsing its across-seed variance: raw ratio $30.8\times$ at $L{=}512$, robust variance tests $p=0.013$--$0.029$, worst seed 86\% vs.\ 14\% (\S\ref{sec:main160}). The variance reduction replicates at 410M with $n{=}4$ per arm (SD 6.5\% vs.\ 16.5\% at $L{=}2048$, Levene $p=0.040$; worst seed 29.4\% vs.\ 18.8\%) (\S\ref{sec:scale410}).
\item \textbf{The active ingredient is anti-periodicity, isolated by controls and made causal by ablation.} Two matched control pretraining arms ($n{=}6$ each) dissect the mechanism: an aperiodic ladder covering the identical low-frequency band (no holonomy class) reproduces none of the reliability effect (variance ratio $1.0\times$ vs.\ standard), and a periodic ladder (integer harmonics, holonomy $+1$) only a fraction ($3.1\times$, n.s.); against $30.8\times$ for holonomy $-1$, this is a dose--response ordering in the boundary condition. An evaluation-time ablation on the trained hybrid models (all weights frozen, only the frequency table swapped back to standard) collapses retrieval from $90.3\%$ to $41.7\%$ at $L{=}512$, with damage concentrated on far-from-query needles ($-55$pp at depth 0.1 vs.\ $-36$pp at depth 0.9): trained models causally depend on the M\"obius geometry, with the damage pattern implicating its long-range structure (\S\ref{sec:controls}).
\item \textbf{A reliability--quality trade-off across positional-encoding families.} A NoPE arm ($n{=}6$; no positional encoding) achieves the best short-context retrieval reliability of all arms ($95.2{\pm}4.0\%$ at $L{=}512$, worst seed 88.8\%), but pays a $13\%$ perplexity tax (33.66 vs.\ 29.66--29.72), redevelops a seed lottery at the full training length (SD 24.1\% at $L{=}2048$ vs.\ hybrid's 8.2\%), and extrapolates worst ($5.6\times$ ppl blow-up at $4\Nmax$). Across the 30 models of the five-arm cohort, retrieval and language-modeling quality anti-align ($r{=}{+}0.43$ between ppl and NIAH, $p{=}0.017$, pooled across arms); hybrid M\"obius is the only arm that combines standard-level perplexity with a high, low-variance retrieval floor (\S\ref{sec:controls}).
\item \textbf{A capacity--scale interaction in allocating the anti-periodic budget.} Splitting the M\"obius heads across multiple scales (``ladder'', $N \in \{512, 1024, 2048\}$) is significantly worse than a single scale at 160M ($-30$pp, $p=0.015$; dipole dilution); yet at 410M ($n{=}4$ per arm) the ordering reverses: ladder attains the best mean NIAH at every length with the smallest seed-variance among the $n{=}4$ arms, and the best extrapolation perplexity ($\ppl$@4096 35.7 vs.\ 38.7, Welch $p=0.006$; all four seed-paired deltas negative, paired $p=0.013$). Concentrate the budget when heads are scarce; distribute it when heads are plentiful (\S\ref{sec:ladder}).
\item \textbf{A characterized failure boundary.} Beyond the training length, all variants collapse to 0\% retrieval through the same critical transition (median retrieved-token rank jumps from ${\sim}2.5$--$6$ to ${\sim}8{,}400$--$9{,}500$ between $L{=}2048$ and $3072$; needle attention mass evaporates $0.073 \to 0.004$ at 3072, $0.003$ by 4096) (\S\ref{sec:collapse}). M\"obius RoPE is for \emph{reliable retrieval within the training window}, not a long-context method (\S\ref{sec:discussion}).
\end{enumerate}

\paragraph{Practical framing.} The intervention is a one-line frequency change on 25\% of heads with no cost at the training length: \textbf{pretraining insurance} that converts a capability that emerges stochastically into one that emerges reliably.

\section{Related Work}

\paragraph{Rotary positional encoding.} RoPE \citep{su2024roformer} encodes relative position by rotating query/key pairs in $d_h/2$ two-dimensional planes at geometrically spaced frequencies $\theta_i = b^{-2i/d_h}$ (base $b = 10^4$). Because the $\theta_i$ share no short common period, the encoding is aperiodic in practice: no shift of realistic length returns the phase configuration to itself, and long-range relative phases are effectively scrambled. All of our comparisons use this standard parameterization as the baseline.

\paragraph{Length generalization for RoPE.} A large body of work modifies RoPE's frequency spectrum to extend usable context: position interpolation \citep{chen2023pi}, NTK-aware scaling and YaRN \citep{peng2024yarn}, LongRoPE \citep{ding2024longrope}, FoPE's Fourier-series spectrum repair \citep{hua2025fope}, and MrRoPE's mixed-radix generalization \citep{tian2026mrrope}. Our goal is orthogonal: we do not extend context. We change the \emph{boundary condition} at a fixed training length, and show explicitly (\S\ref{sec:collapse}) that our method does not help, and slightly hurts, beyond the training window. Among non-rotary families, we pretrain a NoPE arm \citep{kazemnejad2023nope} under the identical protocol as a reference point outside the RoPE family (\S\ref{sec:controls}); ALiBi \citep{press2022alibi} is left to future work. All within-family comparisons are against standard RoPE, the dominant choice in current pretrained models.

\paragraph{Periodic positional encodings.} The closest relative is P-RoPE \citep{huo2026prope}, which pairs sliding-window attention with a periodic positional structure (positions are reset modulo the window, so the encoding returns to itself after one traversal, holonomy $+1$ in our terms). Anti-periodic frequencies are a topologically distinct class (half-integer harmonics, holonomy $-1$; within the class of $N$-(anti)periodic encodings the two holonomy classes are separated, though generic non-periodic ladders interpolate between them), and the sign matters: holonomy $+1$ aliases the two ends of the sequence, whereas $-1$ couples them with a deterministic sign flip; this is the distinction between a cylinder and a M\"obius band. To our knowledge no prior positional encoding uses anti-periodic boundary conditions (prior-art search: July 2026).

\paragraph{M\"obius structures in attention.} M\"obiusAttention \citep{halacheva2024mobiusattention} introduces M\"obius transformations (complex fractional-linear maps) into the attention computation, acting on learned query/weight representations to enrich their geometry. Despite the shared name, it is unrelated to our construction: it modifies the attention nonlinearity with learned projective maps, whereas we modify the \emph{positional boundary condition} with a fixed, parameter-free frequency ladder. Our ``M\"obius'' refers to the $-1$ holonomy of the position bundle, i.e., the topology of the M\"obius band.

\paragraph{Retrieval evaluation and emergence.} Passkey retrieval \citep{mohtashami2023landmark}, popularized as needle-in-a-haystack testing \citep{kamradt2023niah}, is commonly applied to pretrained models at long context; we use it instead as a probe of \emph{whether the capability emerges at all} during small-scale pretraining, across seeds. The instability we document connects to induction heads \citep{olsson2022induction}; our results suggest that at small scale the formation of such retrieval circuitry is seed-dependent. Our intervention can be read as seeding that circuit geometrically.

\paragraph{Physics antecedents.} Our ladder is exactly the Matsubara mode structure of fermionic fields on a thermal circle ($\omega_n = \pi(2n+1)/\beta$; \citealp{matsubara1955new}; the fermionic anti-periodicity condition itself: \citealp{ezawa1957quantum}), with inverse temperature replaced by context length; we import the boundary condition, not any dynamical content.

\section{Theory: Anti-Periodic Frequencies and the Dipole Kernel}
\label{sec:theory}

\subsection{Setup}

RoPE acts on a head of dimension $d_h$ (even) by partitioning it into $P = d_h/2$ planes and rotating plane $i$ of the query/key at position $m$ by angle $m\theta_i$:
\begin{equation}
R(m) = \bigoplus_{i=0}^{P-1} \begin{pmatrix} \cos m\theta_i & -\sin m\theta_i \\ \sin m\theta_i & \cos m\theta_i \end{pmatrix}, \qquad q_m = R(m)\, q, \quad k_n = R(n)\, k,
\end{equation}
so that $\langle q_m, k_n \rangle$ depends only on the relative offset $m-n$. Standard RoPE takes $\theta_i^{\text{std}} = b^{-2i/d_h}$ with $b = 10^4$.

\begin{definition}[M\"obius RoPE]
Fix $N$ (in all experiments, the training context length $\Nmax = 2048$). M\"obius RoPE uses the anti-periodic (half-integer) frequency ladder
\begin{equation}
\boxed{\;\theta_i = \frac{\pi\,(2i+1)}{N}, \qquad i = 0, 1, \dots, P-1.\;}
\end{equation}
\end{definition}

$N$ is fixed at model creation and does not change with the runtime sequence length; extrapolation behavior is governed by the $2N$-periodicity below.

\subsection{Holonomy $-1$: one traversal flips the sign}

After a relative offset of exactly $N$ positions, plane $i$ has rotated by $N\theta_i = \pi(2i+1)$, an odd multiple of $\pi$, hence each planar rotation equals $-I_2$ and
\begin{equation}
R(N) = -I_{d_h}, \qquad R(2N) = +I_{d_h}.
\end{equation}
Each rotation plane realizes a flat bundle over the circle with holonomy $-1$, the M\"obius sign; we use the name for this boundary sign, not as a claim about the topology of the full rank-$d_h$ bundle. Every linear positional feature built from these frequencies (any fixed linear combination of the per-plane sines and cosines) is anti-periodic, $f(m+N) = -f(m)$, and $2N$-periodic; periodic constructions ($R(N) = +I$) cannot distinguish position $m$ from $m+N$, while the anti-periodic class distinguishes them by sign.

\subsection{Closed-form attention signature: a Dirichlet kernel}

Consider the pure positional geometry: identical, per-plane-normalized query and key ($q = k$, equal energy in every plane). The position-only attention coefficient at offset $m$ is the average phase coherence
\begin{equation}
c(m) \;=\; \frac{1}{P} \sum_{i=0}^{P-1} \cos\!\big(m\,\theta_i\big) \;=\; \frac{1}{P} \sum_{i=0}^{P-1} \cos\!\big((2i+1)\,x\big), \qquad x = \frac{\pi m}{N}.
\end{equation}
The half-integer harmonic sum has a classical closed form (geometric series: $\sum_{i=0}^{P-1} \cos((2i+1)x) = \frac{\sin(2Px)}{2\sin x}$); writing $D = d_h = 2P$:
\begin{equation}
\boxed{\;c(m) \;=\; \frac{\sin\!\big(D\,\pi m / N\big)}{D\,\sin\!\big(\pi m / N\big)}\;}
\label{eq:dirichlet}
\end{equation}
--- a (shifted) \textbf{Dirichlet kernel}. This closed form makes four falsifiable predictions:

\begin{itemize}
\item \textbf{(P1) Local lobe.} $c(0) = 1$ with a main lobe of half-width $N/D$: positions within ${\sim}N/D$ of each other cohere, giving ordinary local positional discrimination.
\item \textbf{(P2) Silence in the bulk.} For $m$ away from $0$ and $N$, $|c(m)| \le 1/(D \sin(\pi m/N)) = O(1/D)$: the middle of the sequence contributes at most a few percent.
\item \textbf{(P3) Antipodal dipole.} As $m \to N$, $x \to \pi$ and $c \to -1$: the far end of the context is coupled to the origin with coefficient ${\approx}-1$. The kernel is a \textbf{dipole}: $+1$ lobe at offset 0, $-1$ lobe at offset $N$, near-zero in between; under this diagnostic ($q=k$, equal per-plane energy), a key at one end contributes a deterministic pre-softmax coefficient of ${\approx}{-}1$ at the other end. The sign is benign: the coupling enters pre-softmax, where a sign flip of the learned key projection converts inhibition to excitation.
\item \textbf{(P4) Bounded, $2N$-periodic extrapolation.} For any runtime length $L$, $c$ is exactly given by the same formula; $|c| \le 1$ always, $c(m+N) = -c(m)$, $c(m+2N) = c(m)$. Standard RoPE trivially satisfies $|c|\le1$ as well, but admits no analogous closed form or short exact period: its coefficient at fixed offset drifts quasi-randomly, with no $2N$-periodic structure.
\end{itemize}

\subsection{Numerical verification}
\label{sec:numerics}

All four predictions were tested directly (Phases 1--2.5 on CPU/MPS; Stage A on H100):

\begin{itemize}
\item \textbf{Attention-map signature (Phase 1).} $N \times N$ attention maps under M\"obius frequencies match the Dirichlet closed form with maximum absolute error $1.33 \times 10^{-6}$ (5/5 registered predictions pass), including the sign-flipped anti-diagonal corner (Fig.~\ref{fig:phase1}, appendix).
\item \textbf{End-to-end signal transport (Phase 2; Fig.~\ref{fig:phase2}).} In a single attention layer at $N = 10{,}000$, a value vector planted at position 0 is transported, sign-inverted, to position $N{-}1$ with $\cos(o, V) = -0.9606$: one-hop propagation across ten thousand tokens through the negative dipole lobe. Implementation cost is unchanged (only the constant cos/sin table differs; measured runtime versus sequence length matches the standard-RoPE implementation, log--log slope 1.002).
\item \textbf{Dipole profile census (Phase 2.5; Fig.~\ref{fig:census}).} Full profile scan at $N = 10{,}000$, $D = 32$ confirms a silent bulk (largest middle-50\% response $=$ 4.2\% of the endpoint response, matching the closed form's 4.2\%; 2.2\% at $D=64$) with extremal negative response exactly at the far end (argmin at position 0 as predicted); 3/3 pass.
\item \textbf{Extrapolation (Stage A; Fig.~\ref{fig:stageA}).} With $\theta_i$ frozen at $\Nmax = 2048$, the measured endpoint coefficient $c(L)$ at the ten registered runtime lengths $L \in \{0.25, \dots, 4\} \times \Nmax$ (up to 8192) matches theory with per-point error $\le 1.8 \times 10^{-6}$ (max over the full 800-point float32 sweep: $1.2\times 10^{-5}$); it is bounded and $2\Nmax$-periodic (e.g., $c(L{-}1) = -0.9984$ at $L = 2048$ and $+0.0022$ at $L = 2560$, i.e., offsets 2047 and 2559), while the standard-RoPE coefficient varies erratically ($+0.20$, $+0.24$, $+0.014$, $-0.011$, \dots) with no closed form. All four registered judgments pass.
\end{itemize}

\paragraph{Why might a dipole help retrieval?} Retrieval circuits must learn to move information from an arbitrary earlier position to the current query; under standard RoPE the long-range positional geometry is unstructured, and whether training discovers such a circuit appears to be seed-dependent (\S\ref{sec:main160}). The M\"obius dipole provides the network with a \emph{deterministic long-range channel at initialization}; the hypothesis tested below is that this scaffold makes retrieval-circuit formation reliable rather than left to chance.

\section{Method: Hybrid and Ladder Head Allocation}
\label{sec:method}

Anti-periodicity trades bulk positional resolution for boundary coupling, so imposing it on all heads is likely harmful for language modeling. We therefore use mixed allocations:

\paragraph{Hybrid (main method).} A fraction $\rho = 0.25$ of attention heads (deterministically the first $\lceil \rho H \rceil$ heads of every layer) use M\"obius frequencies with a single $\Nmax = 2048$ equal to the training context; the remaining 75\% use standard RoPE ($b = 10^4$). For the 160M model ($H = 12$) this is 3 M\"obius heads per layer; for 410M ($H = 16$), 4 heads. No new parameters; the only change is the cos/sin table applied to those heads. FLOPs, memory, and throughput are identical to the baseline.

\paragraph{Ladder (ablation).} To test whether multi-scale dipoles help, the ladder variant assigns 3 heads per layer M\"obius frequencies with different scales $\Nmax \in \{512, 1024, 2048\}$ (one head per scale, a positional ``skip list'' of dipoles), the rest standard. This probes whether several weaker boundary couplings at staggered ranges beat one strong coupling at the full context scale. The ladder always uses 3 M\"obius heads: at 160M this matches hybrid's budget (3/12), but at 410M it is 3/16 ($\rho{=}18.75\%$) vs.\ hybrid's 4/16 ($\rho{=}25\%$), so the 410M ladder-vs-hybrid comparison confounds allocation with a slightly smaller budget.

\paragraph{Controls (mechanism dissection).} Anti-periodicity confounds two ingredients: where the frequencies live (a narrow low-frequency band) and how they are quantized (half-integer harmonics, holonomy $-1$). Two matched control allocations separate them, each identical to hybrid except for the special heads' frequency table: \textbf{band} uses a geometric, mutually incommensurate ladder spanning exactly the M\"obius band $[\pi/N,\, \pi(d_h{-}1)/N]$: same band placement, no holonomy class ($R(N) \neq \pm I$); \textbf{periodic} uses integer harmonics $\theta_i = 2\pi(i{+}1)/N$ (the nearest integer-harmonic band, $[2\pi/N,\,\pi d_h/N]$, one frequency bin from the M\"obius band), with holonomy $+1$ (the cylinder to M\"obius's band). Together with hybrid (holonomy $-1$) they dissect band placement from boundary condition across three boundary classes.

\paragraph{NoPE and mobius (all-heads).} A NoPE arm (no positional encoding on any head) is pretrained under the identical protocol as an out-of-family reference (\S\ref{sec:controls}); all-heads mobius is used only in the synthetic suite (\S\ref{sec:synthetic}).

\section{Experiments}
\label{sec:experiments}

All pretraining uses one pipeline: decoder-only transformers (GPT-NeoX tokenizer), FineWeb-Edu \citep{penedo2024fineweb} streamed and packed, sequence length 2048, 2.0B tokens (3814 steps $\times$ 524{,}288 tokens), AdamW, lr $6\times 10^{-4}$ cosine, bf16, H100. Two pythia-like scales \citep{biderman2023pythia}: 160M (12 layers, $d_{\text{model}}{=}768$, 12 heads) and 410M-class (405M parameters; 24 layers, $d_{\text{model}}{=}1024$, 16 heads); $d_h = 64$ for both. Same-seed runs within a training round share identical data order, so same-round within-seed comparisons are paired (the control arms used a mirrored data-serving path; limitation 9); seed-paired tests corroborate the independent-samples tests reported below (e.g., paired $t$ at $L{=}512$, hybrid vs.\ standard: $p=0.110$, vs.\ Welch $p=0.090$).

\paragraph{Evaluation.} (i) Validation perplexity at 2048 and extrapolated lengths 4096/8192 (training-time validation losses and these evaluation perplexities are computed on different held-out samples of the stream and are not numerically interchangeable). (ii) NIAH passkey retrieval: the needle sentence ``Remember this important fact: the secret pass-key is \emph{word}.'' is planted at depth $d \in \{0.1, \dots, 0.9\}$ inside coherent English filler and queried at the end (``Question: what is the secret pass-key? Answer: the secret pass-key is''); 32 trials per depth (16 words $\times$ 2 contexts) $\times$ 5 depths $=$ 160 trials per length per seed, $L \in \{128, \dots, 8192\}$; top-1 accuracy, plus the correct token's rank in the model's output distribution (0 = top-1) recorded for every trial. \emph{Harness note:} an initial harness with random packed-token filler returned a spurious 0\% for every model; all results use the corrected harness (see \S\ref{sec:discussion}, limitation 6).

\subsection{Synthetic convergence}
\label{sec:synthetic}

100 training runs (25 per positional variant: none / standard / mobius / hybrid; mixed associative-recall, multi-needle, and reverse-copy tasks; small softmax-attention models) measure learning efficiency rather than ceiling. Final accuracy is indistinguishable across RoPE variants (65.5--65.6\%), but M\"obius reaches the 90\% accuracy threshold in 28\% fewer steps than standard (350 vs.\ 487; hybrid 403; Table~\ref{tab:stageB}, appendix); an earlier three-seed local pilot at $N=256$ shows the same convergence ordering (Fig.~\ref{fig:pilot}). This is consistent with the scaffold hypothesis: the dipole does not add capacity but accelerates the formation of recall circuits.

\subsection{Main result (160M pretraining, $n = 6$ seeds): retrieval becomes reliable}
\label{sec:main160}

\begin{figure}[t]
\centering
\includegraphics[width=0.82\linewidth]{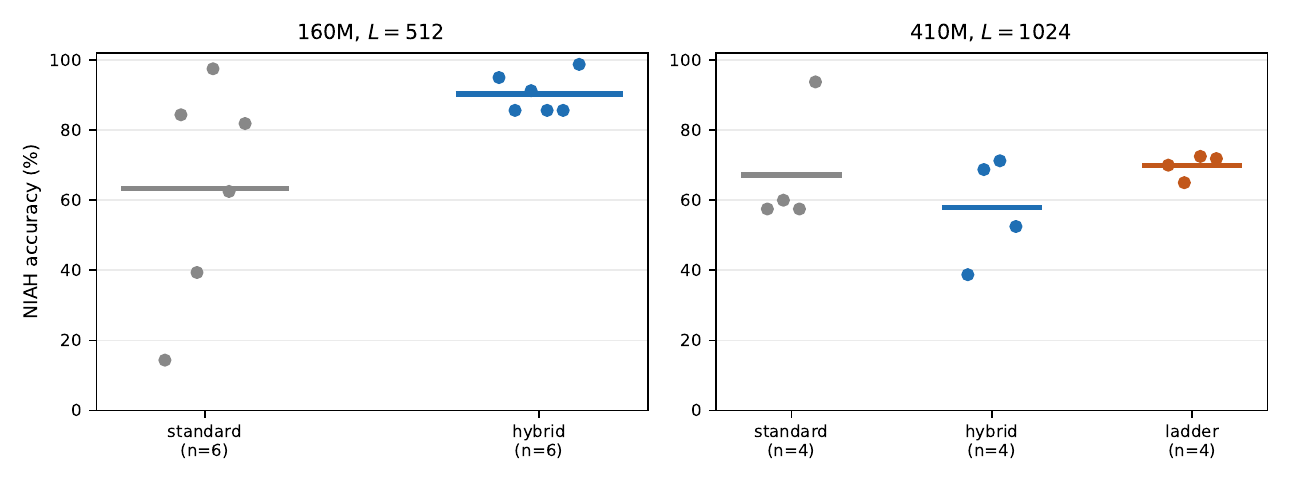}
\caption{Seed-level NIAH scatter (regenerated from archived raw evaluations). \emph{Left:} the 160M retrieval seed lottery at $L{=}512$ ($n{=}6$ per arm): standard RoPE spans 14--98\% while hybrid clusters at 86--99\%. \emph{Right:} at 410M ($L{=}1024$, $n{=}4$ per arm) the reliability property transfers to the multi-scale ladder variant.}
\label{fig:scatter}
\end{figure}

Six seeds per condition (hybrid vs.\ standard), otherwise identical. Perplexity at the training length is at parity: $29.66\pm0.09$ vs.\ $29.72\pm0.24$, $p = 0.63$; final validation losses are likewise statistically indistinguishable ($3.333\pm0.018$ vs.\ $3.333\pm0.020$). NIAH separates the conditions sharply (Table~\ref{tab:main160}; Figs.~\ref{fig:scatter} and \ref{fig:rniah}; loss parity in Fig.~\ref{fig:valloss}).

\begin{table}[t]
\centering
\resizebox{\textwidth}{!}{%
\begin{tabular}{lccccc}
\toprule
$L$ & hybrid & standard & $\Delta$ (pp) & Welch $p$ & Cohen's $d$ \\
\midrule
128 & 67.8 $\pm$ 21.9\% (46\%) & 46.0 $\pm$ 30.9\% (6\%) & $+21.8$ & 0.192 & $+0.81$ \\
256 & 90.0 $\pm$ 8.0\% (82\%) & 65.2 $\pm$ 30.7\% (22\%) & $+24.8$ & 0.107 & $+1.10$ \\
512 & \textbf{90.3 $\pm$ 5.7\% (86\%)} & \textbf{63.3 $\pm$ 31.4\% (14\%)} & $+27.0$ & 0.090 & $+1.20$ \\
1024 & 55.0 $\pm$ 15.1\% (38\%) & 32.9 $\pm$ 25.6\% (6\%) & $+22.1$ & 0.106 & $+1.05$ \\
2048 & 27.4 $\pm$ 8.2\% (18\%) & 17.9 $\pm$ 12.0\% (5\%) & $+9.5$ & 0.146 & $+0.92$ \\
\bottomrule
\end{tabular}}
\caption{160M NIAH accuracy, mean $\pm$ SD over 6 seeds (worst seed in parentheses). No mean difference reaches $p<0.05$ at $n{=}6$; the reliability evidence lives in the variance tests (Table~\ref{tab:variance}).}
\label{tab:main160}
\end{table}

\paragraph{Variance is the central finding.} The per-seed spread is more informative than the means. Standard RoPE seeds at $L{=}512$: 14, 84, 39, 98, 62, 82\% (a lottery). Hybrid seeds: 95, 86, 91, 86, 86, 99\%. The robust statistic is the \emph{worst-seed floor}: 86\% vs.\ 14\%; the hybrid's worst seed beats the standard arm's mean. The raw variance ratio at $L{=}512$ is $30.8\times$ (one-sided F $p=0.0009$) and survives distribution-robust tests ($p=0.013$--$0.029$; Table~\ref{tab:variance}). Part of the raw ratio is a ceiling effect (logit-scale ratio $3.4\times$, $p=0.10$), but the scale-free facts stand: hybrid's SD (5.7\%) sits near the iid-binomial reference floor (${\approx}2.3$--$3.8\%$ at 160 trials; trials share words and contexts, so the true noise floor may be larger) while standard's (31.4\%) is ${\sim}8\times$ above it. In deployment terms, within our sample: standard seeds ranged from 14\% to 98\% while every hybrid seed retrieved at or above 86\%, though six seeds cannot bound the population worst case.

\begin{table}[t]
\centering
\small
\setlength{\tabcolsep}{4pt}
\begin{tabular}{lcccccc}
\toprule
$L$ & SD hybrid & SD standard & var ratio & F-test $p$ & worst hyb & worst std \\
\midrule
128 & 21.9\% & 30.9\% & 2.0$\times$ & 0.235 & 46\% & 6\% \\
256 & 8.0\% & 30.7\% & 14.7$\times$ & 0.0052 & 82\% & 22\% \\
\textbf{512} & \textbf{5.7\%} & \textbf{31.4\%} & \textbf{30.8$\times$} & \textbf{0.0009} & \textbf{86\%} & \textbf{14\%} \\
1024 & 15.1\% & 25.6\% & 2.9$\times$ & 0.135 & 38\% & 6\% \\
2048 & 8.2\% & 12.0\% & 2.1$\times$ & 0.213 & 18\% & 5\% \\
\bottomrule
\end{tabular}
\caption{Across-seed variance of 160M NIAH accuracy ($n{=}6$; all in-window lengths shown). F-test $p$ is one-sided; at $L{=}512$ the robust tests remain significant (Levene 0.014, Brown--Forsythe 0.029, Fligner 0.013).}
\label{tab:variance}
\end{table}

\paragraph{Retrieval failure is invisible to perplexity.} Across these 12 models, Pearson correlation between $\ppl$@2048 and NIAH@512 is $+0.206$ ($\approx 0$; if anything the sign is opposite to a ``bad seeds are worse language models'' explanation). The standard seed with 14\% retrieval has $\ppl$ 29.59, third-best in the entire 12-model cohort. Retrieval and language modeling are decoupled skills at this scale (indeed, across the 30-model five-arm cohort the association runs significantly against quality; \S\ref{sec:controls}), which is exactly why the seed lottery is a practical risk: standard training telemetry will not catch it.

\paragraph{Where the advantage lives.} Depth-resolved accuracy (Fig.~\ref{fig:heatmap}) shows the hybrid advantage across all depths at $L \le 512$ ($+18$ to $+36$pp), concentrating at late depths at $L{=}2048$ ($+29$pp at $d = 0.9$); we report depth patterns descriptively: the kernel magnitude is symmetric in $m \leftrightarrow N{-}m$, so single-lobe geometric readings of depth are not warranted. Hybrid wins on 16/16 passkey words.

\paragraph{Robustness of the finding itself.} Three internal checks argue the effect is not an evaluation artifact: an independent 2-seed pilot with bootstrap CIs on pooled trials found the same gaps ($L{=}512$: $+38.8$pp, 95\% CI $[+29.4, +47.5]$, 5000 resamples); every hybrid seed sits above the standard arm's mean at $L{=}512$; and the effect has structure noise would not produce (systematic depth structure, monotone rank separation, uniform per-word wins). The failure mode is clear: standard seeds 0 and 2 retrieve 14.4\% and 39.4\% at $L{=}512$ with healthy perplexities (29.59 and 29.73; the cohort's best is 29.42, seed 5). Two of the six standard seeds retrieve below 40\% at $L{=}512$, and nothing in training telemetry flags them.

\subsection{Isolating the active ingredient: controls, a causal ablation, and the NoPE trade-off}
\label{sec:controls}

Three follow-up experiment families ($n{=}6$ seeds each, protocol identical to \S\ref{sec:main160}) dissect why hybrid works.

\paragraph{Band vs.\ boundary condition.} Table~\ref{tab:controls} compares five arms at $L{=}512$. The band control (same low-frequency band as M\"obius, but aperiodic) is statistically indistinguishable from standard RoPE in mean and variance (ratio $1.0\times$): concentrating frequencies at long wavelengths does nothing by itself. The periodic control (holonomy $+1$) recovers a small, non-significant fraction of the reliability ($3.1\times$, Levene $p=0.15$). Only holonomy $-1$ produces the effect ($30.8\times$, all robust tests significant). The ordering (aperiodic $1.0\times$ $<$ periodic $3.1\times$ $<$ anti-periodic $30.8\times$) is a dose--response in the boundary condition, and the controls also fall below hybrid in mean (periodic: $-22.0$pp, $p=0.028$; band: $-30.8$pp, $p=0.058$, marginal). Perplexity is at parity for all RoPE-family arms (Table~\ref{tab:controls}), so the controls are quality-neutral; within this design, the reliability difference tracks the boundary condition (with the caveats that the periodic band sits one frequency bin away and that limitation 9 notes a serving-path difference).

\begin{table}[t]
\centering
\footnotesize
\setlength{\tabcolsep}{4pt}
\resizebox{\textwidth}{!}{%
\begin{tabular}{lcccccc}
\toprule
arm ($n{=}6$) & holonomy & NIAH@512 & worst & var-ratio & Lev./BF/Flig. & $\ppl$@2048 \\
\midrule
standard & --- (aperiodic) & 63.3 $\pm$ 31.4\% & 14.4\% & 1$\times$ & --- & 29.72 $\pm$ 0.24 \\
band ctrl & --- (aperiodic) & 59.5 $\pm$ 30.9\% & 19.4\% & 1.0$\times$ & .83/.85/.62 & 29.65 $\pm$ 0.10 \\
periodic ctrl & $+1$ & 68.3 $\pm$ 17.8\% & 37.5\% & 3.1$\times$ & .15/.19/.09 & 29.69 $\pm$ 0.06 \\
\textbf{hybrid} & $\mathbf{-1}$ & \textbf{90.3 $\pm$ 5.7\%} & \textbf{85.6\%} & \textbf{30.8}$\times$ & \textbf{.014/.029/.013} & 29.66 $\pm$ 0.09 \\
\midrule
NoPE & (no PE) & 95.2 $\pm$ 4.0\% & 88.8\% & 61.4$\times$ & .009/.019/.006 & \textbf{33.66 $\pm$ 0.10} \\
\bottomrule
\end{tabular}}
\caption{Mechanism dissection at 160M ($L{=}512$; variance ratios and robust variance tests vs.\ standard). Band placement alone reproduces nothing; holonomy $+1$ a fraction; holonomy $-1$ the full effect at perplexity parity. NoPE is most reliable here but pays a 13\% perplexity tax (bold = anomaly).}
\label{tab:controls}
\end{table}

\paragraph{Causal ablation: trained models need the geometry.} We take the six trained hybrid models and, at evaluation time only, (i) swap the M\"obius heads' frequency table back to standard RoPE with all weights frozen (\textbf{rope-swap}), (ii) zero the M\"obius heads' output projection (\textbf{zero-mob}), or (iii) zero an equal number of standard heads (\textbf{zero-std}, damage control). Rope-swap, which changes no parameters, collapses retrieval from $90.3{\pm}5.7\%$ to $41.7{\pm}26.9\%$ at $L{=}512$ (and $90.0 \to 28.1\%$ at 256, $55.0 \to 32.8\%$ at 1024), on every seed. Depth-resolved, the damage concentrates on needles far from the query ($-55.2$ and $-58.9$pp at depths 0.1 and 0.3, easing to $-36.5$pp at depth 0.9): the trained circuit depends most on the removed geometry in the far half of the context, where query--needle distances are longest, consistent with the M\"obius planes' low-frequency long-range structure (of which the endpoint dipole is the extreme case). This establishes causal dependence on the anti-periodic geometry; attributing it to the endpoint lobe specifically rests on the closed form and this distance pattern, and remains partly interpretive. We caution that the head-zeroing conditions are less diagnostic: zero-std collapses retrieval even harder ($7.2\%$ at 512), i.e., removing any 3 of 12 heads from these small models is catastrophic, so the causal weight rests on rope-swap, which removes nothing but the geometry.

\paragraph{Task-scope probe: the reliability effect is specific to single-needle retrieval.} To test generality beyond passkey NIAH, we evaluated all 30 archived checkpoints on two harder retrieval families under the identical protocol (trial-paired across models; 160 trials per length each): \textbf{KV-lookup} (4--12 key--value pairs planted in filler; query one key's value) and \textbf{multi-needle} (three named passkeys; query one by name). Both tasks sit near the capability floor at this scale (arm means 2.5--25.6\% at $L \in \{512, 1024\}$), and no arm, hybrid included, reproduces the variance-collapse signature there (all Levene tests n.s.; hybrid attains the best RoPE-family means on both tasks, $+4.5$pp KV and $+8.4$pp multi-needle at $L{=}512$, but n.s.\ at $n{=}6$; NoPE's mean advantage partially generalizes, $+10.4$pp multi-needle, $p=0.026$). Two readings fit these results: floor compression limits variance comparisons (the mirror image of the NIAH ceiling caveat), and, consistent with the mechanism, the dipole is a positional end-to-end channel, not a content-addressing device: it reliably transports a single planted signal across the context, but provides neither key$\to$value binding (KV) nor disambiguation among same-format competitors (multi-needle). The reliability claim of this paper is therefore scoped to the single-needle regime that the geometry actually addresses.

\paragraph{The NoPE surprise, and the trade-off.} The out-of-family NoPE arm is the most reliable short-context retriever in the study ($95.2{\pm}4.0\%$ at $L{=}512$; flat depth profile 92.7--96.9\%, consistent with purely content-based addressing), evidence that the seed lottery is largely \emph{induced by positional structure interfering with retrieval-circuit formation}. However, this reliability comes at a cost on every other axis: (i) it pays a 13\% perplexity tax at the training length (33.66 vs.\ 29.66--29.72; final val loss 3.474 vs.\ 3.333), (ii) its reliability is confined to short/mid context: at $L{=}2048$ its own lottery re-emerges (SD 24.1\% vs.\ hybrid's 8.2\%), and (iii) it extrapolates worst of all arms ($\ppl$ ratio $5.60\times$ at $L{=}8192$ vs.\ standard's $3.89\times$). Across the 30 models of the five-arm control cohort (ladder excluded) the pattern generalizes: perplexity and NIAH@512 are positively correlated ($r{=}{+}0.431$, $p{=}0.017$; a pooled, between-arm association, as within-arm samples are too small to establish a model-level trade-off), i.e., worse language models retrieve more reliably. This exposes a trade-off surface across positional-encoding families on which hybrid M\"obius is the only arm that pays neither cost: standard-RoPE perplexity and a near-NoPE reliability floor within the training window (band matches the perplexity without the reliability; NoPE the reliability without the perplexity).

\subsection{Scaling to 410M-class models (405M parameters)}
\label{sec:scale410}

Round 2 repeats the comparison at 405M parameters/2B tokens (seeds 0--1; all three arms were later extended to $n{=}4$, seeds 0--3; for the M\"obius arms see \S\ref{sec:ladder}). Perplexity again at parity, hybrid marginally ahead: 25.16 $\pm$ 0.06 vs.\ 25.26 $\pm$ 0.12 (both $n{=}4$). NIAH at the training length: 38.1 $\pm$ 6.5\% (hybrid, $n{=}4$) vs.\ 34.4 $\pm$ 16.5\% (standard, $n{=}4$); per-seed 38.8/39.4/29.4/45.0\% vs.\ 41.2/23.1/18.8/54.4\%. Two observations:

\begin{enumerate}
\item \textbf{The robustness effect persists at scale, now with statistical support.} Standard RoPE still draws unstable seeds at 410M (range 18.8--54.4\% at $L{=}2048$); hybrid's four seeds span 29.4--45.0\%. At matched $n{=}4$, the variance reduction is significant (SD 6.5\% vs.\ 16.5\%, Levene $p=0.040$) and the worst-seed floor rises from 18.8\% to 29.4\%. (At 410M the strongest reliability signature belongs to the ladder variant, \S\ref{sec:ladder}.)
\item \textbf{The mean-accuracy gap shrinks.} The 160M advantage of $+22$ to $+27$pp at $L \le 1024$ becomes a deficit-to-parity at 410M. The driver is the baseline: standard RoPE's absolute retrieval improves markedly with capacity ($L{=}512$: $63 \to 88\%$ at $n{=}4$) while hybrid is roughly flat. With more capacity, standard training finds retrieval circuits on its own more often: the dipole scaffold matters most exactly where emergence is most fragile.
\end{enumerate}

Depth-resolved accuracy (see also Fig.~\ref{fig:distance}, 160M cohort) sharpens the picture: at $L{=}2048$ all variants score 0\% for needles in the first third of the context, and the hybrid advantage is confined to the latest depth ($+23.4$pp at $d{=}0.9$; $+0.8$pp at $d{=}0.7$); as above, we treat depth profiles as descriptive. The picture is not uniformly favorable: at $L{=}512$ hybrid trails standard at most depths ($-18.8$pp at $d{=}0.1$, $-11.7$pp at $d{=}0.9$, parity only at mid depths), consistent with its overall mean deficit at this length (Table~\ref{tab:ladder410}). Sub-top-1 analysis mirrors 160M: median rank at $L{=}2048$ is 2.5 (hybrid) vs.\ 6 (standard), pooled over the $n{=}4$ seeds.

\subsection{Ladder ablation: a negative result at 160M, and a confirmed reversal at 410M}
\label{sec:ladder}

\paragraph{160M, $n = 6$: multi-scale is clearly worse.} The ladder variant (M\"obius heads split across $\Nmax \in \{512, 1024, 2048\}$) was hypothesized to add mid-range retrieval structure. It does the opposite (Table~\ref{tab:ladder160}).

\begin{table}[t]
\centering
\resizebox{\textwidth}{!}{%
\begin{tabular}{lccccc}
\toprule
$L$ & ladder & hybrid & standard & $\Delta$(lad$-$hyb, pp) & Welch $p$ \\
\midrule
512 & 60.3 $\pm$ 20.7\% & 90.3 $\pm$ 5.7\% & 63.3 $\pm$ 31.4\% & $\mathbf{-30.0}$ & \textbf{0.015} \\
1024 & 25.8 $\pm$ 15.0\% & 55.0 $\pm$ 15.1\% & 32.9 $\pm$ 25.6\% & $\mathbf{-29.2}$ & \textbf{0.007} \\
2048 & 13.5 $\pm$ 9.0\% & 27.4 $\pm$ 8.2\% & 17.9 $\pm$ 12.0\% & $\mathbf{-13.9}$ & \textbf{0.020} \\
\bottomrule
\end{tabular}}
\caption{160M ladder ablation ($n{=}6$): multi-scale allocation is significantly worse than single-scale hybrid.}
\label{tab:ladder160}
\end{table}

Ladder is statistically indistinguishable from standard and far below hybrid; the effective retrieval window (mean $\ge 50\%$) is 512 tokens for ladder and standard vs.\ 1024 for hybrid. Interpretation: with only 3 M\"obius heads per layer, splitting across scales leaves one head per scale (\textbf{dipole dilution}). This negative result is primary evidence for the single-dipole mechanism.

\paragraph{410M, $n = 4$ per M\"obius arm: the reversal, confirmed.} At 410M the picture flips. An initial $n{=}2$ signal (ladder best on every metric, but with a data-order confound from mid-training stream-crash resumes) prompted four additional clean runs (seeds 2--3 for both ladder and hybrid, identical protocol and single-run data order). The reversal's direction survives (Table~\ref{tab:ladder410}), with an important refinement about which part is statistically solid.

\begin{table}[t]
\centering
\small
\setlength{\tabcolsep}{4pt}
\resizebox{\textwidth}{!}{%
\begin{tabular}{lcccc}
\toprule
metric & \textbf{ladder ($n{=}4$)} & hybrid ($n{=}4$) & standard ($n{=}4$) & lad.\ vs.\ hyb.\ $p$ \\
\midrule
NIAH @256 & \textbf{81.9 $\pm$ 9.8\%} & 69.7 $\pm$ 20.2\% & 72.0 $\pm$ 19.2\% & 0.33 \\
NIAH @512 & \textbf{90.3 $\pm$ 5.0\%} & 80.9 $\pm$ 13.5\% & 88.3 $\pm$ 12.3\% & 0.27 \\
NIAH @1024 & \textbf{69.8 $\pm$ 3.4\%} & 57.8 $\pm$ 15.2\% & 67.2 $\pm$ 17.7\% & 0.21 \\
NIAH @2048 & \textbf{38.9 $\pm$ 3.0\%} & 38.1 $\pm$ 6.5\% & 34.4 $\pm$ 16.5\% & 0.84 \\
$\ppl$ @2048 & \textbf{24.70 $\pm$ 0.69} & 25.16 $\pm$ 0.06 & 25.26 $\pm$ 0.12 & 0.28 \\
$\ppl$ @4096 & \textbf{35.72 $\pm$ 0.63} & 38.71 $\pm$ 1.12 & 36.65 $\pm$ 1.08 & $\mathbf{0.006}$ \\
$\ppl$ @8192 & \textbf{82.92 $\pm$ 3.02} & 96.48 $\pm$ 3.82 & 85.21 $\pm$ 5.38 & $\mathbf{0.002}$ \\
val loss & 3.1456 $\pm$ 0.0308 & 3.1509 $\pm$ 0.0234 & 3.1475 $\pm$ 0.0364$^{\dagger}$ & --- \\
\bottomrule
\end{tabular}}
\caption{410M three-way comparison, now $n{=}4$ per arm (standard seeds 2--3 added in Round 4). Welch tests, ladder vs.\ hybrid. $^{\dagger}$val loss for standard from the original $n{=}2$ logs.}
\label{tab:ladder410}
\end{table}

We note three findings. \textbf{(i) The extrapolation-perplexity advantage is the solid one:} ladder beats hybrid on $\ppl$@4096 in all four seed-paired comparisons (deltas $-3.46$, $-3.09$, $-4.01$, $-1.41$; Welch $p = 0.006$ at 4096, $0.002$ at 8192; paired $t$: $0.013$, $0.005$), eliminating the data-order confound: multi-scale M\"obius pays no extrapolation tax and extrapolates best of all variants including standard. \textbf{(ii) The NIAH advantage is real in means but driven by variance:} ladder has the highest mean at every length and the smallest seed-variance of all three arms ($\pm3.4$ vs.\ hybrid's $\pm15.2$ and standard's $\pm17.7$ at $L{=}1024$, all $n{=}4$), but the mean differences do not individually reach significance ($p = 0.21$--$0.84$) and per-pair results are heterogeneous. The correct claim is that at 410M ladder inherits the reliability property that hybrid exhibits at 160M. \textbf{(iii) The 160M$\to$410M sign flip itself is unambiguous} (from $-30$pp, $p{=}0.015$ against ladder to $+9$--$12$pp in its favor at $L\le1024$, $+0.8$pp at 2048; formal capacity$\times$allocation interaction at $L{=}512$: $z=3.5$, $p=0.0005$): a capacity--scale interaction (concentrate the budget when heads are scarce, distribute it when heads are plentiful), with the \S\ref{sec:method} caveat that at 410M ladder also carries a smaller budget (18.75\% vs.\ 25\%), so allocation and budget are not fully separable.

\subsection{Beyond the training length: a critical collapse, common to all variants}
\label{sec:collapse}

Retrieval collapses past the training window in every arm: NIAH is exactly 0.0\% for hybrid, standard, and ladder at $L \in \{3072, 4096, 6144, 8192\}$ at both scales, every seed; the control arms follow (band and periodic: exactly 0\% at $L{=}3072$; NoPE: arm mean 2.5\% at 3072, one seed retaining 11.2\%, exactly 0\% at $L\ge4096$; NoPE additionally suffers the worst extrapolation perplexity, $5.60\times$ at $L{=}8192$). Three probes show this is a sharp transition, not gradual decay:

\begin{itemize}
\item \textbf{Rank cliff} (see also Fig.~\ref{fig:ranks} for the 160M cohort). Median rank of the correct token (410M, pooled over the $n{=}4$ seeds): 2.5 (hybrid) / 6 (standard) / 3 (ladder) at $L{=}2048$ $\to$ $8{,}436$ / $9{,}547$ / $8{,}672$ at $L{=}3072$ (chance ${\approx} 25{,}000$). The fraction of trials with rank $\le 10$ falls from ${\sim}52$--$55\%$ to $\le 0.3\%$. Even sub-top-1 retrieval signal vanishes across the boundary.
\item \textbf{Attention-mass evaporation} (Fig.~\ref{fig:collapse}). Layer-wise probing (160M, seed 2): peak attention mass on the needle span is 0.073 (hybrid) / 0.043 (standard) at $L{=}2048$, concentrated at layer 9 (the retrieval layer) and collapses to 0.004 / 0.004 at 3072 and 0.003 / 0.003 at 4096, with the peak relocating to layer 0 (i.e., no layer attends to the needle at all).
\item \textbf{Entropy flattening.} Normalized attention entropy at the retrieval layer rises from 0.54 ($L{=}2048$) toward uniform (0.85 at 3072, 0.90 at 4096). Attention does not mis-retrieve out-of-distribution; it diffuses.
\end{itemize}

This is a generic OOD failure of models trained at fixed context, indifferent to the positional variant. M\"obius theory says its geometry stays bounded under extrapolation (\S\ref{sec:numerics}), and it does, but bounded geometry cannot compensate for query/key statistics that were never trained past $\Nmax$.

\paragraph{Extrapolation perplexity cost.} Hybrid pays a real, if modest, tax out-of-window: 160M $\ppl$@4096 55.05 vs.\ 51.19 ($\times$1.86 vs.\ $\times$1.72 over $\ppl$@2048; $p = 0.031$) and @8192 132.65 vs.\ 115.47 ($p < 0.001$); at 410M ($n{=}4$ vs.\ $n{=}4$), $\times$1.54 vs.\ $\times$1.45 at 4096 and 96.5 vs.\ 85.2 at 8192. The cost shrinks with scale but persists. Ladder pays no such tax ($\times$1.45 at 410M, the best extrapolation $\ppl$ of all three 410M arms). Plausibly the $c \approx -1$ endpoint lobe re-entering at $L > \Nmax$ injects spurious long-range coupling into text that no longer warrants it.

\section{Discussion and Limitations}
\label{sec:discussion}

\paragraph{What we claim.} Within the training context length, replacing the frequency table on 25\% of heads with anti-periodic frequencies (i) leaves perplexity unchanged, (ii) increases mean in-context retrieval accuracy at small scale, and (iii), the central finding, collapses the across-seed variance of retrieval capability at 160M (robust variance tests $p=0.013$--$0.029$, unadjusted; limitation 4), raising the observed worst-seed floor from 14\% to 86\%; at 410M ($n{=}4$ per arm) the variance reduction is significant (Levene $p=0.040$) and the floor rises from 19\% to 29\%, with the strongest reliability signature transferring to the ladder allocation. The mechanism is no longer merely consistent with the closed-form dipole; it is \emph{isolated} (matched band and periodic controls reproduce none/little of the effect, \S\ref{sec:controls}) and \emph{causally probed} (freezing all weights and removing only the geometry collapses retrieval, with damage concentrated at long range): a deterministic end-to-end positional channel that suppresses the stochasticity of retrieval-circuit formation.

\paragraph{What we do not claim.} M\"obius RoPE is not a long-context method: it does not extend usable context by a single token, all variants collapse past $\Nmax$ (\S\ref{sec:collapse}), and hybrid slightly worsens extrapolation perplexity. The correct positioning is \emph{in-training-length retrieval robustness}.

\paragraph{Limitations and threats to validity.}
\textbf{(1) Scale:} largest model is 405M / 2B tokens; the 160M$\to$410M trend (baseline improves, mean gap shrinks, variance gap persists) is extrapolation from two points.
\textbf{(2) Task scope (now measured, not merely unknown):} the variance-collapse effect is established for single-needle passkey retrieval; on harder families (KV-lookup, multi-needle) 160M models are near the capability floor and no arm shows the signature (\S\ref{sec:controls}), so claims should not be extrapolated beyond the single-needle regime; QA-over-context and downstream benchmarks remain untested.
\textbf{(3) Ladder NIAH means not individually significant at 410M:} the 160M negative result is solid ($n{=}6$, $p \le 0.02$) and the 410M extrapolation-$\ppl$ advantage consistent ($p=0.006$, all four seed pairs agreeing), but 410M NIAH mean differences do not reach $p<0.05$ at $n{=}4$; that claim rests on consistent ordering plus variance reduction. Practical rule: single-scale hybrid at ${\le}$160M-class, ladder at ${\ge}$410M-class.
\textbf{(4) Statistical power:} at $n{=}6$, no 160M hybrid-vs-standard mean-difference test reaches $p < 0.05$ (min $p = 0.09$); the mean claims are effect-size claims ($d$ up to 1.2), and the inferential weight rests on the robust variance tests and the worst-seed floor; the raw $30.8\times$ ratio is partly a ceiling artifact (logit-scale $3.4\times$). All $p$-values are unadjusted; under a conservative Bonferroni correction over the five evaluated lengths the $L{=}512$ robust variance tests attenuate to $p=0.065$--$0.145$ (and the 410M Levene to $p=0.20$), so single-length significance should be read as exploratory; the cross-length consistency of the variance ordering (Table~\ref{tab:variance}) and the recurrence at 410M carry the inferential weight. All per-seed values for the headline comparisons are reported.
\textbf{(5) Hyperparameters not swept:} $\rho = 0.25$, head placement, and $\Nmax =$ training length were fixed a priori.
\textbf{(6) Evaluation fragility:} our first NIAH harness silently returned 0\% for all models (filler-packing bug); we fixed and re-scored, and report this to flag how brittle retrieval evals are.
\textbf{(7) Controls at one scale:} the band/periodic controls and the NoPE arm were run at 160M only; the boundary-class dissection is not yet replicated at 410M.
\textbf{(8) Head-zeroing ablations are confounded:} removing any 3 of 12 heads is broadly destructive (\S\ref{sec:controls}); the causal claim rests on the weight-preserving rope-swap condition.
\textbf{(9) Data serving path:} the Round-4 arms (band, periodic, NoPE, 410M standard seeds 2--3) were trained on a mirrored copy of the identical FineWeb-Edu shards served from local disk (a CDN outage made Hugging Face Hub streaming unavailable); same data and shuffle procedure, but byte-identical stream order relative to the original arms is not guaranteed. Arm-level comparisons are across seed distributions and are not expected to be sensitive to serving order, but we cannot fully exclude residual data-order effects; we disclose this for completeness.

\paragraph{Outlook.} The seed lottery itself deserves attention: capability variance across seeds is invisible to loss-based monitoring yet operationally decisive. Deterministic geometric scaffolds, of which the anti-periodic dipole is one instance, may be a general tool for converting fragile emergent circuits into reliable ones.

\section{Conclusion}

We introduced anti-periodic boundary conditions into positional encoding: M\"obius RoPE, whose $-1$ holonomy couples the ends of the training context through a closed-form Dirichlet dipole. On 25\% of heads at zero cost, it leaves perplexity untouched while making retrieval reliable: the observed 160M worst-seed floor rises from 14\% to 86\% (robust variance tests $p=0.013$--$0.029$, unadjusted), replicating at 410M (Levene $p=0.040$). Matched controls show the effect is specific to anti-periodicity (same-band aperiodic and periodic ($+1$) ladders reproduce a $1.0\times$ and $3.1\times$ fraction of the $30.8\times$ variance reduction), and a weight-frozen geometry swap collapses trained models' retrieval, establishing causal dependence on the anti-periodic geometry. Across the wider design space, NoPE attains reliability at a 13\% perplexity tax and standard RoPE attains quality at the price of a retrieval lottery; the anti-periodic hybrid is the only observed configuration that avoids both costs. A multi-scale variant fails at 160M yet leads at 410M: a capacity-dependent allocation rule. A one-line frequency swap provides free insurance against the retrieval seed lottery within the training context.

\section*{CRediT authorship contribution statement}
\textbf{Ji Ho Bae:} Conceptualization, Methodology, Software, Validation, Formal analysis, Investigation, Resources, Data curation, Writing -- original draft, Writing -- review \& editing, Visualization, Project administration.

\section*{Declaration of competing interest}
The author declares that he has no known competing financial interests or personal relationships that could have appeared to influence the work reported in this paper.

\section*{Funding}
This research did not receive any specific grant from funding agencies in the public, commercial, or not-for-profit sectors; cloud compute costs were self-funded.

\section*{Data availability}
All raw evaluation records (per-trial retrieval outcomes), training logs, final model weights and optimizer checkpoints for the 48 pretrained models (all later-round models and all surviving original checkpoints), the exact training-data shards (with SHA-256 manifests), the complete training/evaluation/analysis code, and per-run launch commands are archived. Toward the journal's research-data policy (Option C): at submission time this deposit is not yet public; the complete package will be deposited in a public repository with a citable DOI upon acceptance, and is available to the editors and reviewers on request during review. One 410M checkpoint (hybrid, seed 1) was lost to a compute-instance failure after its evaluation completed; its statistics are intact and a same-seed retrained checkpoint is archived. All Welch $p$-values use the exact Student-$t$ distribution with Welch--Satterthwaite degrees of freedom.

\appendix
\section{Additional tables and figures}
\label{app:figures}

\begin{table}[h]
\centering
\begin{tabular}{lccc}
\toprule
variant & mean final acc & runs reaching 90\% & mean steps to 90\% \\
\midrule
none & 59.9\% & 12/25 & 621 \\
standard & 65.5\% & 15/25 & 487 \\
\textbf{hybrid} & 65.6\% & 15/25 & \textbf{403} \\
\textbf{mobius} & 65.6\% & 15/25 & \textbf{350} \\
\bottomrule
\end{tabular}
\caption{Synthetic suite, 100 runs: final accuracy is indistinguishable across RoPE variants, but M\"obius reaches the 90\% threshold in 28\% fewer steps than standard.}
\label{tab:stageB}
\end{table}

\begin{figure}[h]
\centering
\includegraphics[width=0.95\linewidth]{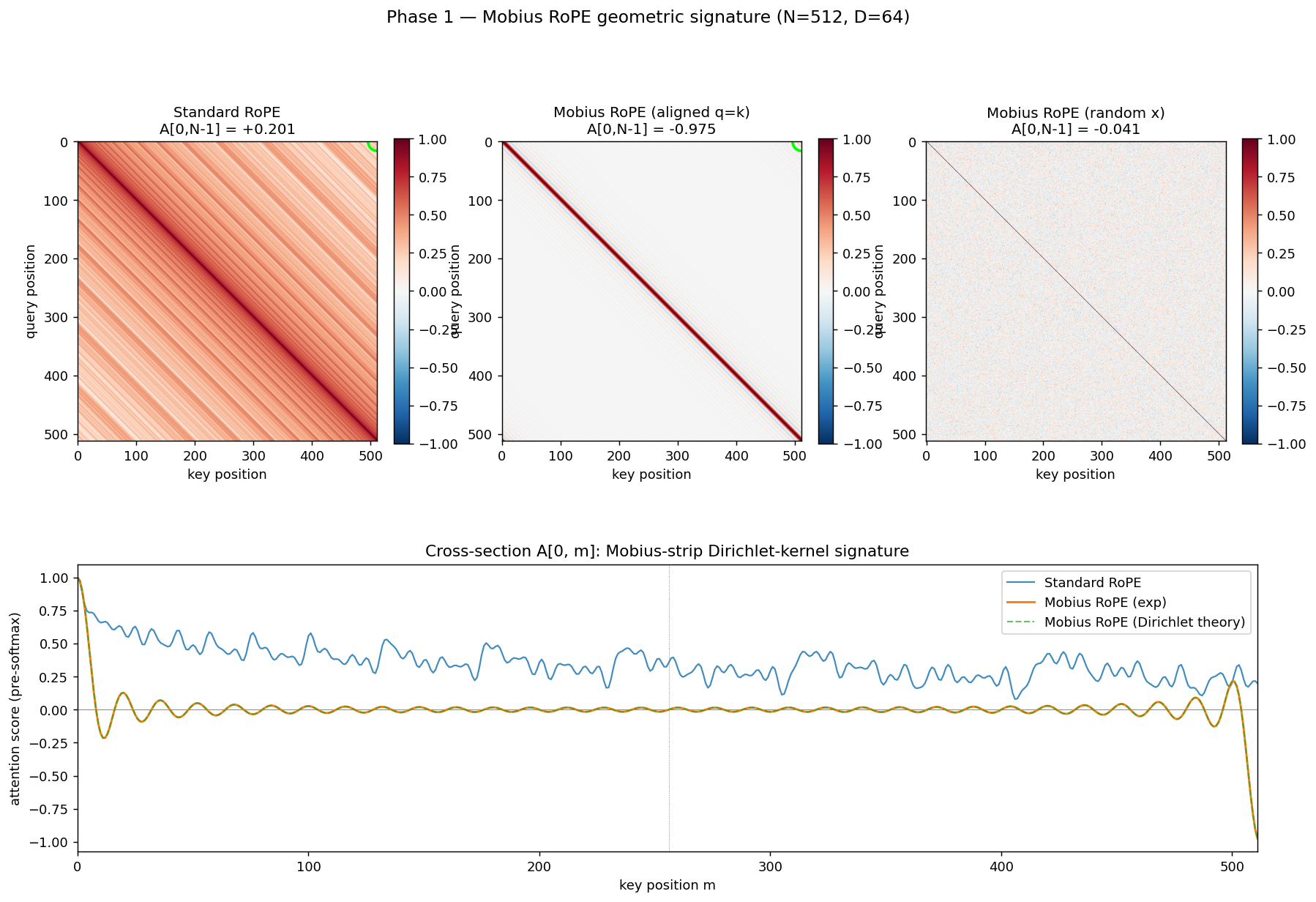}
\caption{Phase 1: M\"obius vs.\ standard attention maps. The M\"obius map exhibits the predicted Dirichlet-kernel structure with the sign-flipped anti-diagonal (end-to-end dipole); maximum deviation from the closed form (Eq.~\ref{eq:dirichlet}) is $1.33\times10^{-6}$.}
\label{fig:phase1}
\end{figure}

\begin{figure}[h]
\centering
\includegraphics[width=0.9\linewidth]{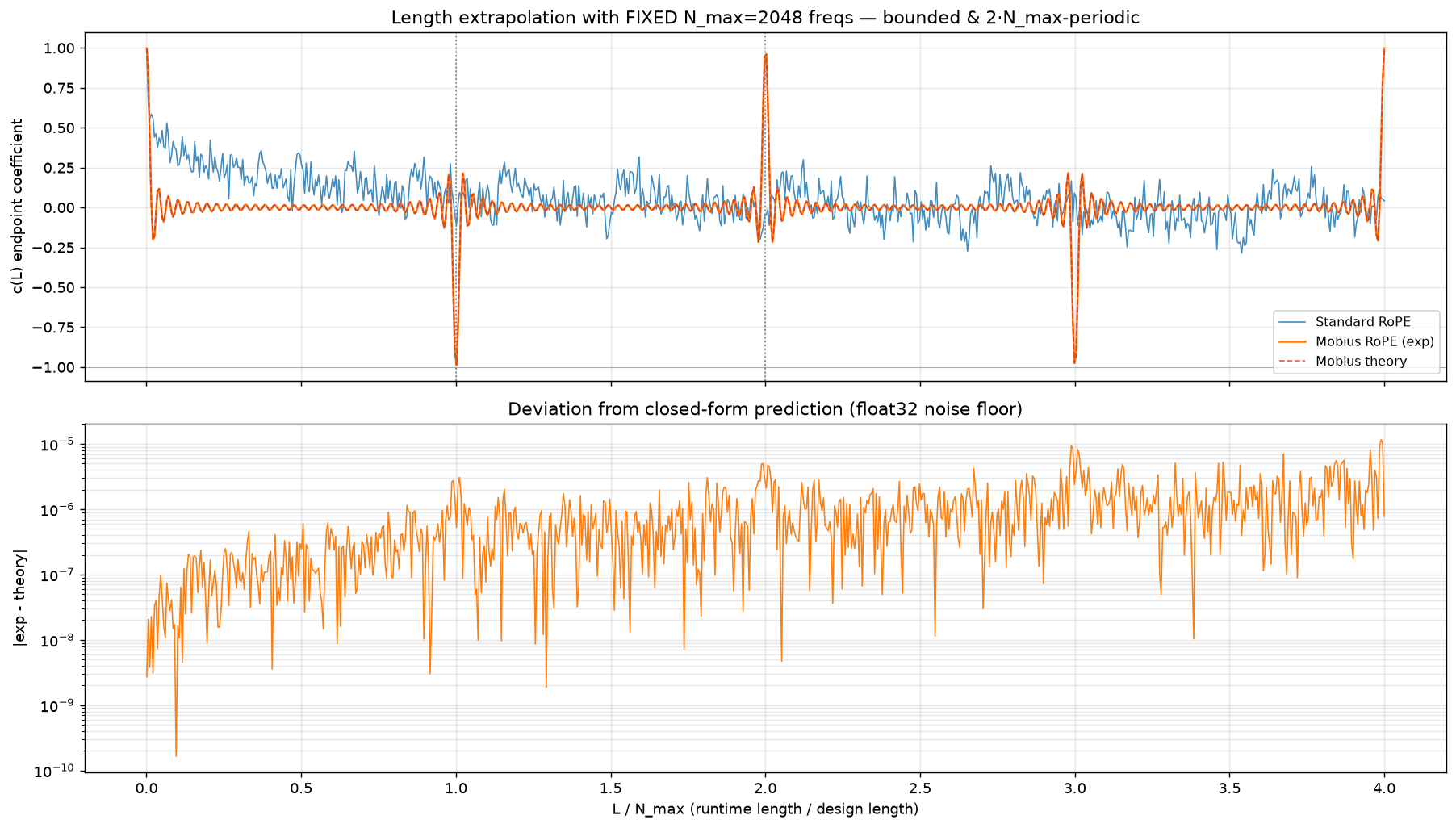}
\caption{Stage A: with frequencies frozen at $\Nmax=2048$, the M\"obius endpoint coefficient is bounded and $2\Nmax$-periodic under extrapolation to $4\times \Nmax$, matching Eq.~\ref{eq:dirichlet} to $\le 1.8\times10^{-6}$ at the ten registered test points (float32 sweep max $1.2\times10^{-5}$, bottom panel), while standard RoPE's coefficient drifts erratically.}
\label{fig:stageA}
\end{figure}

\begin{figure}[h]
\centering
\includegraphics[width=0.9\linewidth]{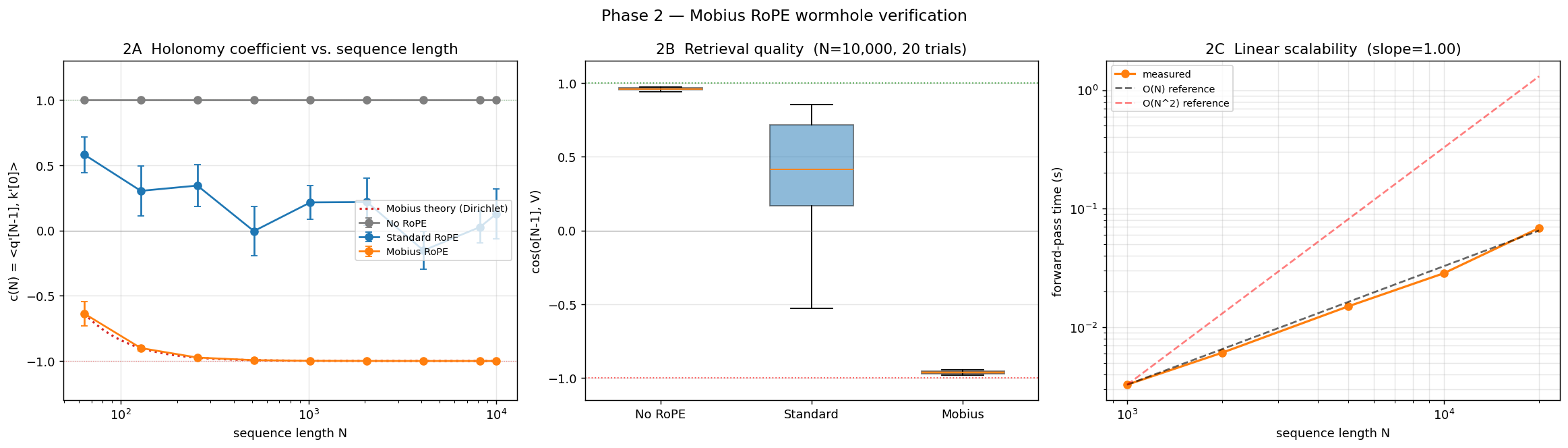}
\caption{Phase 2: end-to-end signal transport at $N=10{,}000$ through the negative dipole lobe ($\cos(o,V)=-0.9606$) and runtime scaling (log--log slope 1.002).}
\label{fig:phase2}
\end{figure}

\begin{figure}[h]
\centering
\includegraphics[width=0.9\linewidth]{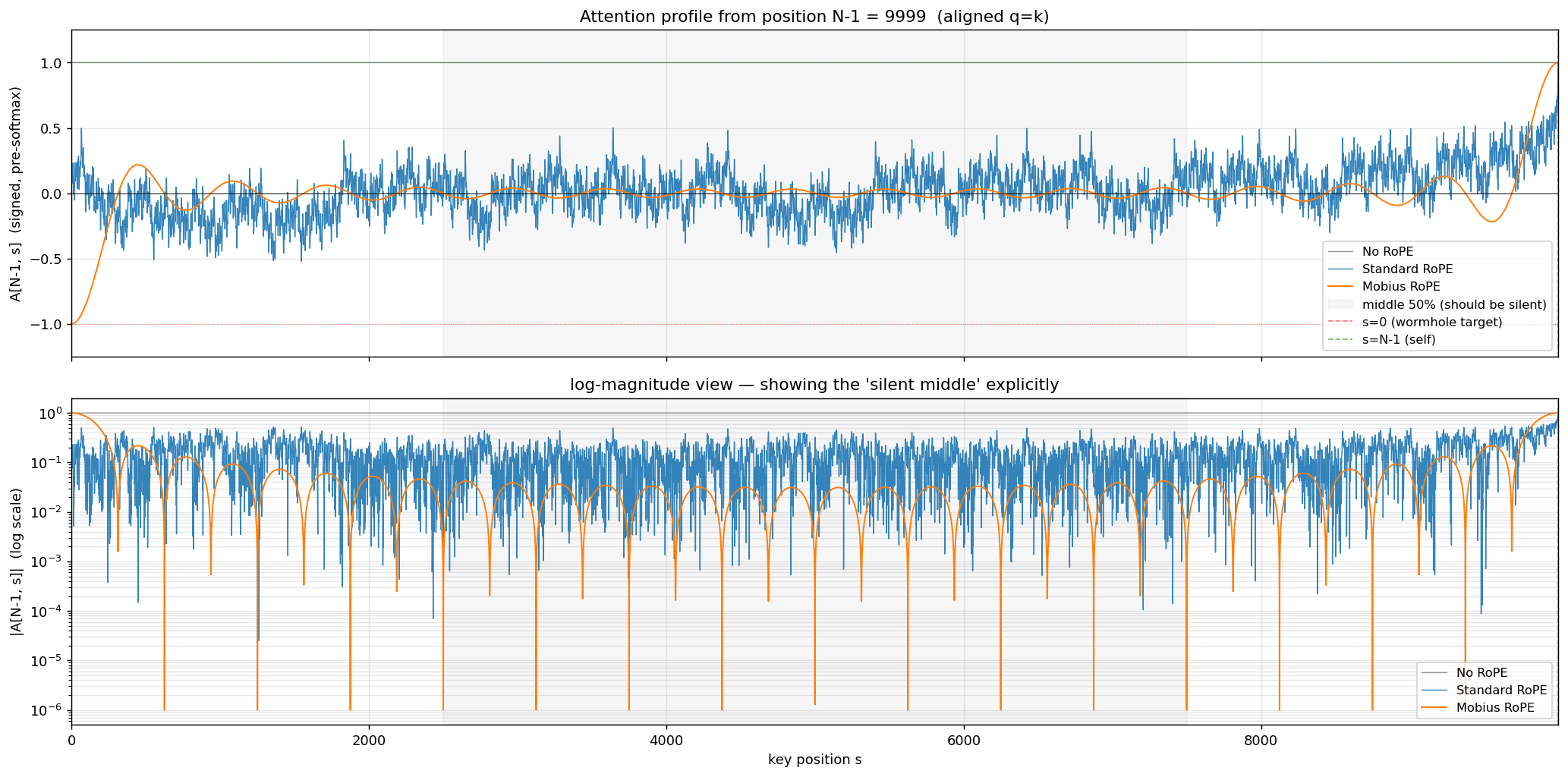}
\caption{Phase 2.5: dipole profile census at $N=10{,}000$, $D=32$: silent bulk (largest middle-50\% response 4.2\% of the endpoint response; 2.2\% at the $d_h=64$ used in pretraining), extremal negative response at the far end.}
\label{fig:census}
\end{figure}

\begin{figure}[h]
\centering
\includegraphics[width=0.85\linewidth]{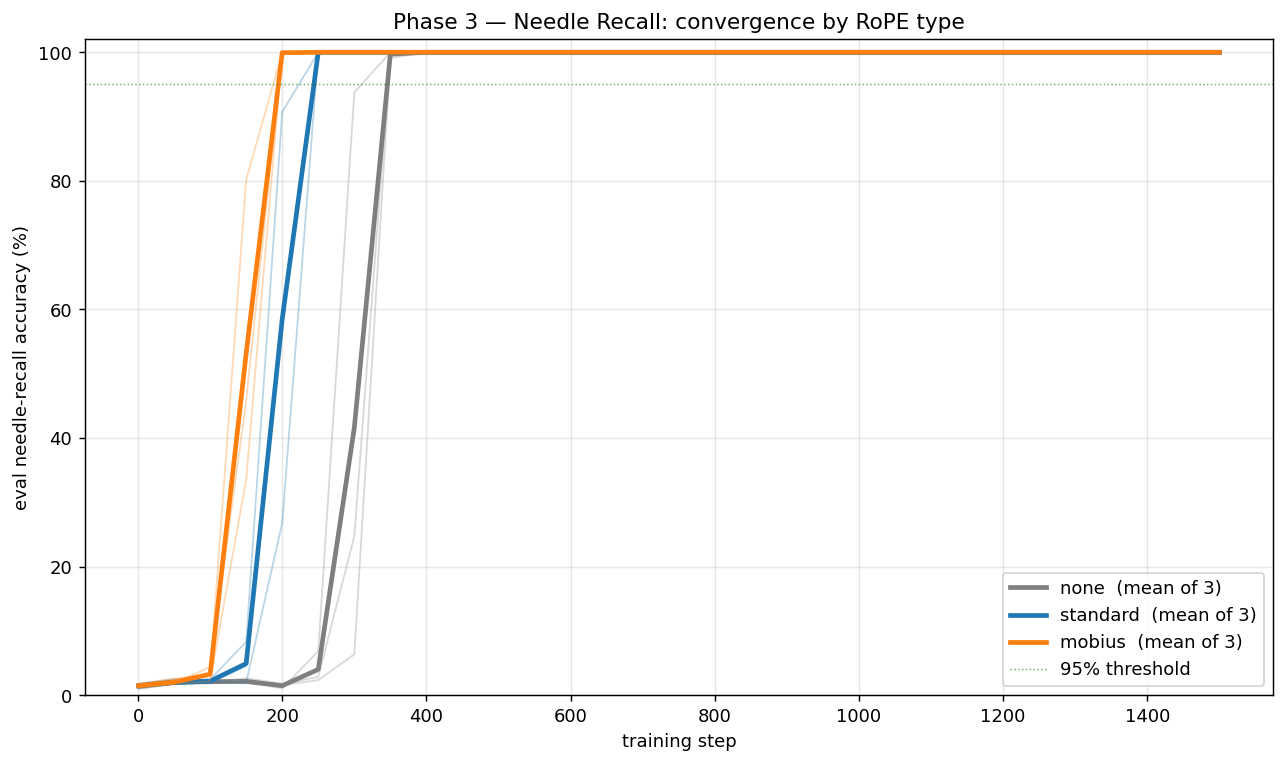}
\caption{Local pilot: needle-recall training curves at $N=256$; M\"obius converges ${\sim}20\%$ faster on 3/3 seeds.}
\label{fig:pilot}
\end{figure}

\begin{figure}[h]
\centering
\includegraphics[width=0.95\linewidth]{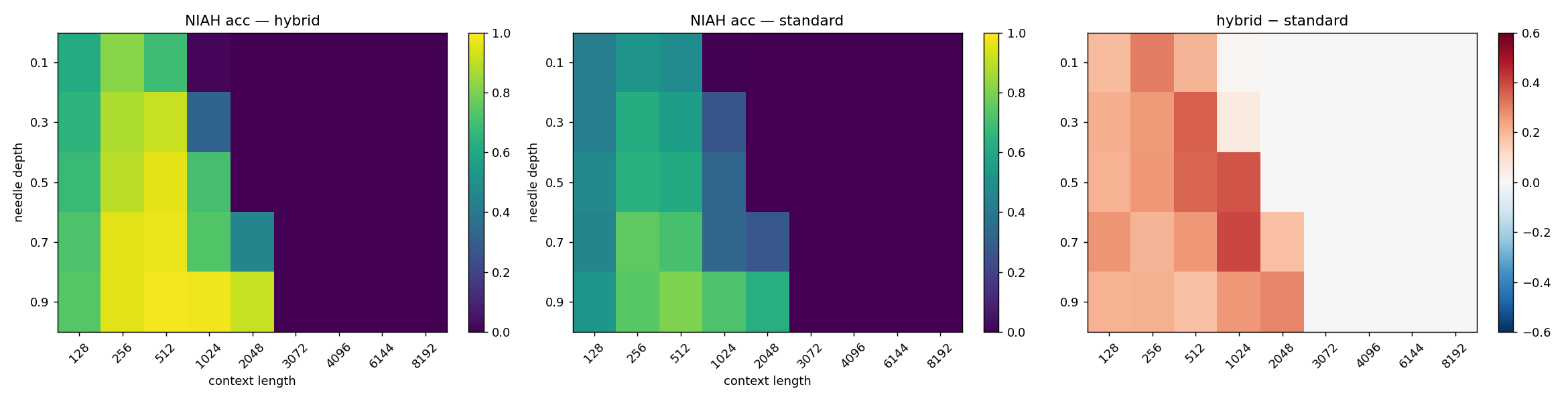}
\caption{160M ($n{=}6$): depth $\times$ length NIAH heatmaps, hybrid vs.\ standard, and their difference.}
\label{fig:heatmap}
\end{figure}

\begin{figure}[h]
\centering
\includegraphics[width=0.9\linewidth]{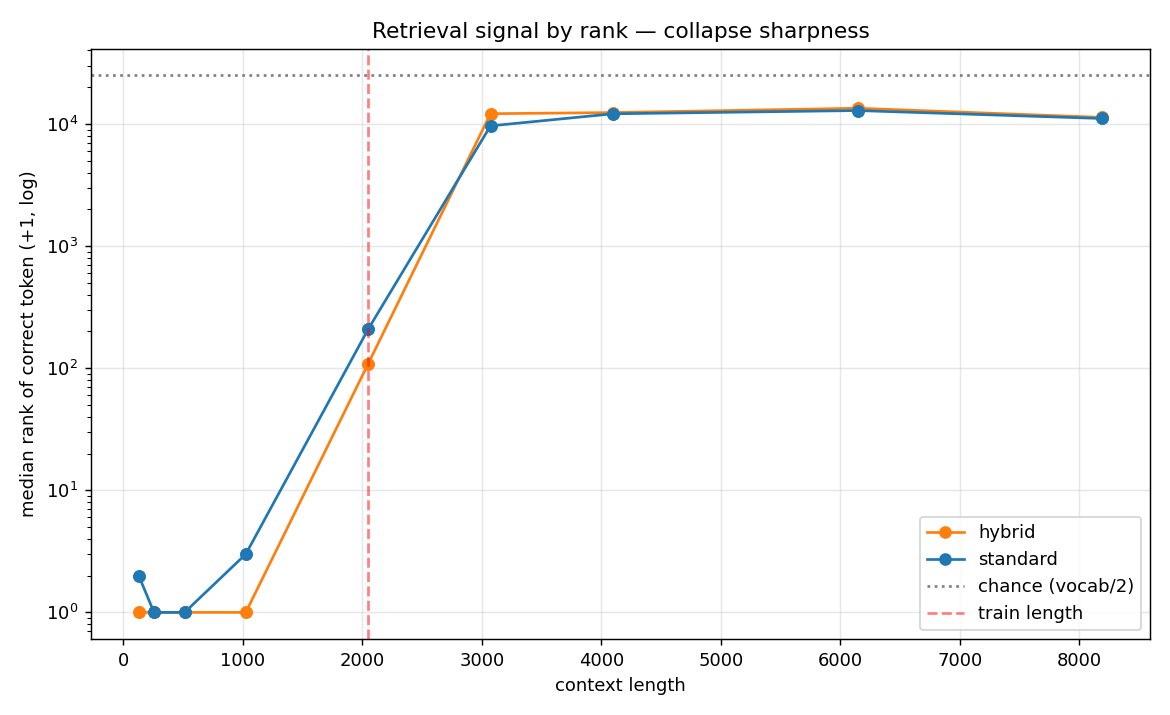}
\caption{Rank of the correct token vs.\ context length (160M cohort, pooled across seeds and depths): sub-top-1 retrieval signal separates hybrid from standard within the training window and vanishes for both beyond it.}
\label{fig:ranks}
\end{figure}

\begin{figure}[h]
\centering
\includegraphics[width=0.9\linewidth]{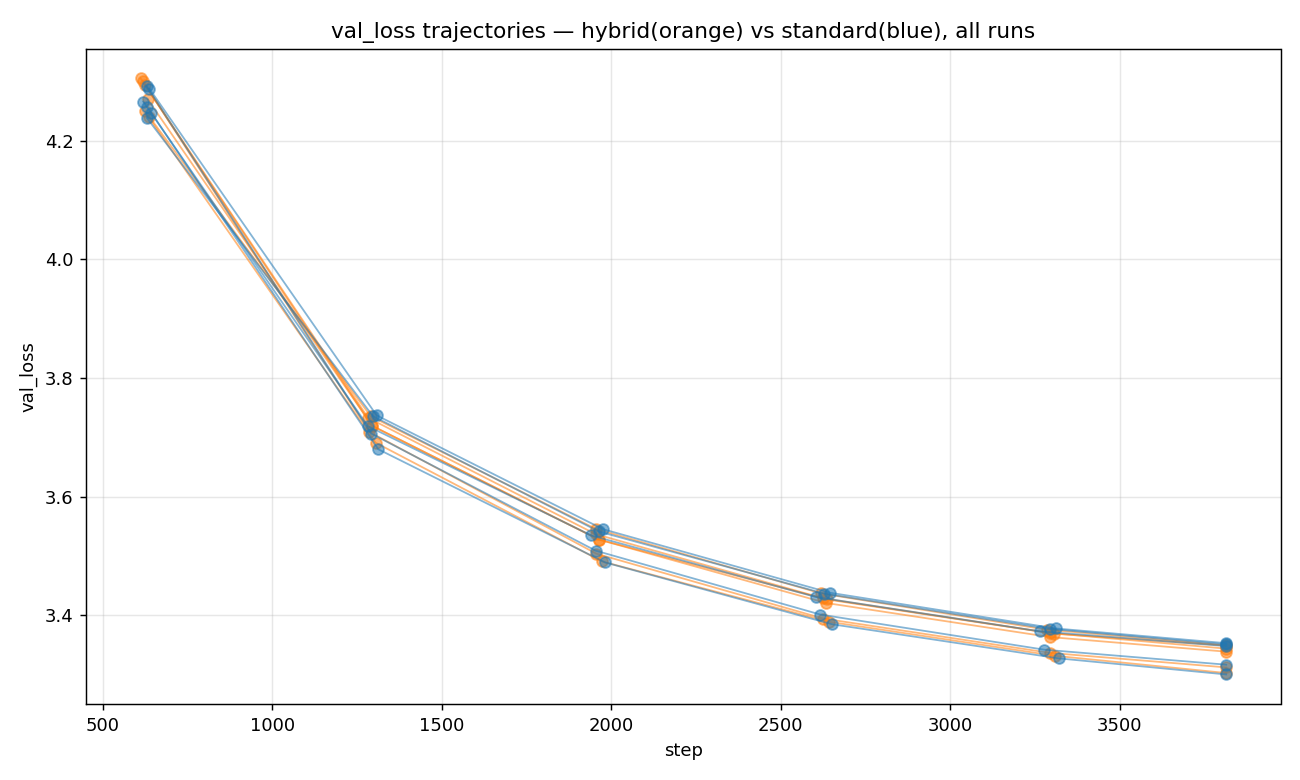}
\caption{Validation-loss trajectories for the 12 original 160M runs: all converge to 3.30--3.35 (loss parity across conditions); the retrieval divergence (Table~\ref{tab:main160}) is invisible here.}
\label{fig:valloss}
\end{figure}

\begin{figure}[h]
\centering
\includegraphics[width=0.9\linewidth]{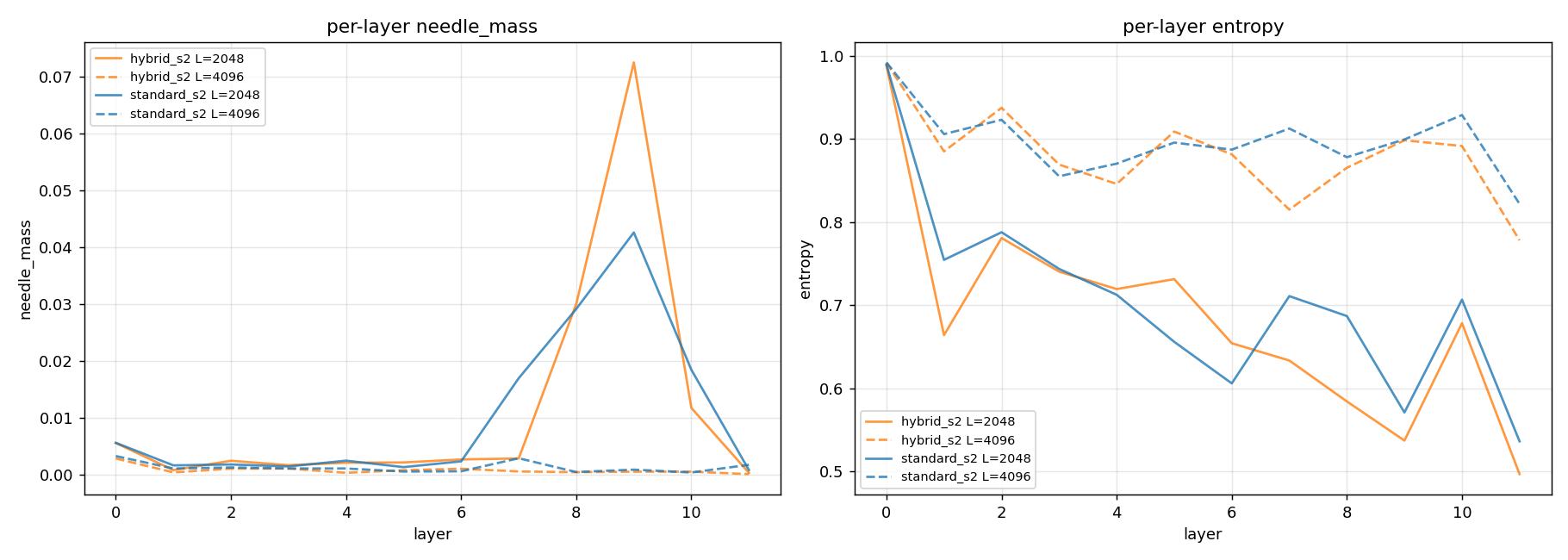}
\caption{Collapse trace (160M, seed 2): per-layer needle attention mass at $L = 2048$ vs.\ $4096$ (left) and normalized attention entropy rising toward uniform (right). The retrieval layer (layer 9) collapses to ${\approx}0.003$--$0.004$ needle mass past the training length; the intermediate $L{=}3072$ values quoted in \S\ref{sec:collapse} come from the same trace data.}
\label{fig:collapse}
\end{figure}

\begin{figure}[h]
\centering
\includegraphics[width=0.9\linewidth]{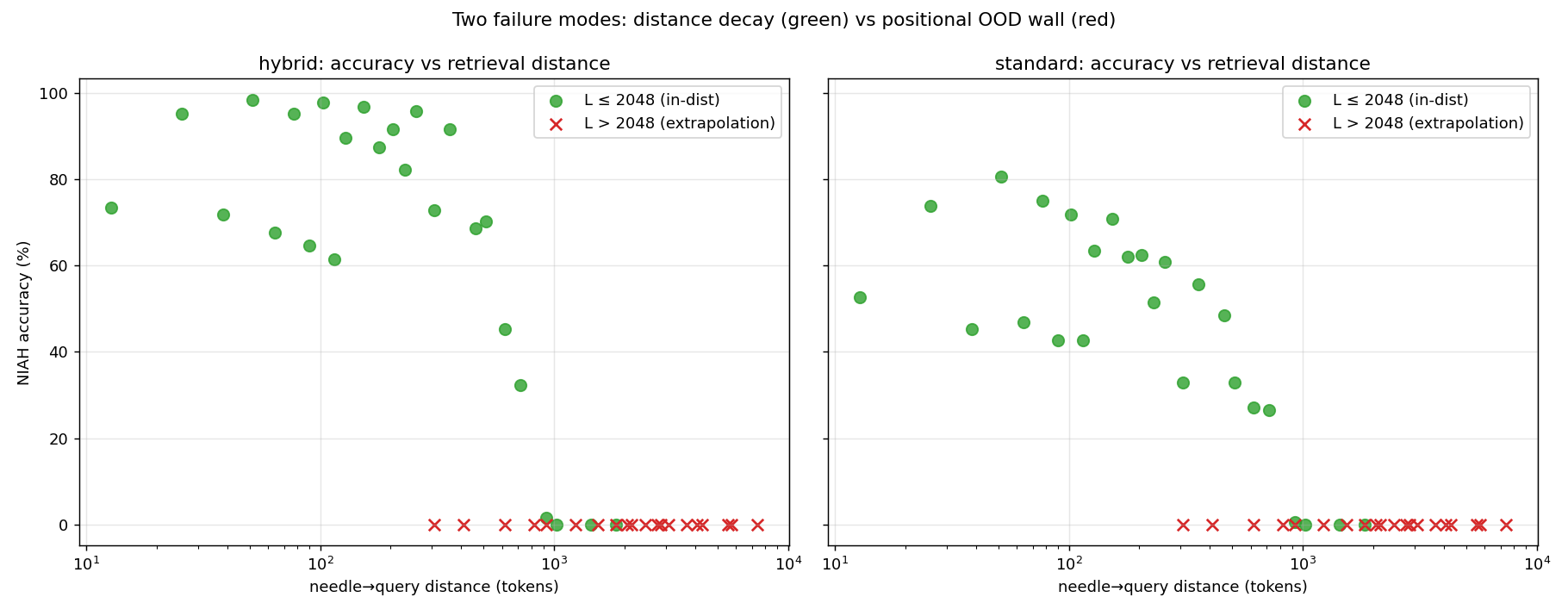}
\caption{Retrieval distance analysis (160M cohort): accuracy as a function of query--needle distance.}
\label{fig:distance}
\end{figure}

\begin{figure}[h]
\centering
\includegraphics[width=0.95\linewidth]{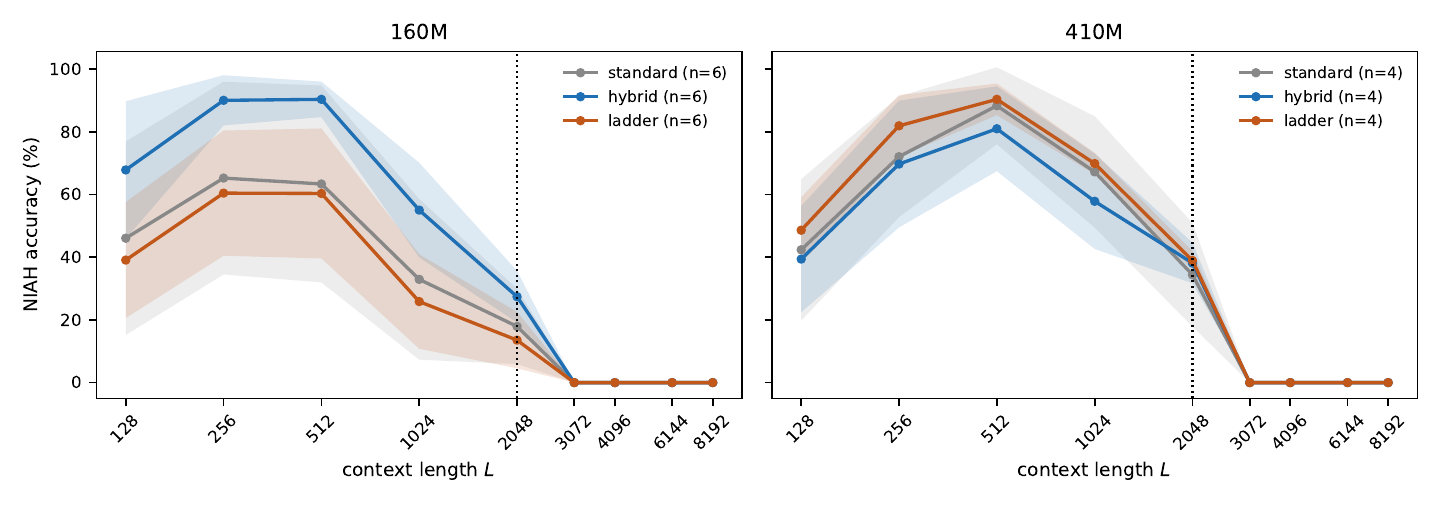}
\caption{NIAH accuracy vs.\ context length (mean $\pm$ SD across seeds; regenerated from archived raw evaluations), standard/hybrid/ladder at both scales, 160M ($n{=}6$ per arm) and 410M ($n{=}4$ per arm, including the extended standard arm), with the training-length boundary marked.}
\label{fig:rniah}
\end{figure}

\bibliographystyle{elsarticle-harv}
\bibliography{references}

\end{document}